\documentclass[runningheads]{llncs}

\usepackage{eccv}

\usepackage{eccvabbrv}

\usepackage{graphicx}
\usepackage{booktabs}
\usepackage{tikz}
\usepackage{svg}

\usepackage{amsmath}
\usepackage{amsfonts}
\usepackage{amssymb}
\usepackage{pifont}% http://ctan.org/pkg/pifont

\usepackage[accsupp]{axessibility}  % Improves PDF readability for those with disabilities.

\newcommand{\cmark}{\ding{51}}%
\newcommand{\xmark}{\ding{55}}%

\usepackage[pagebackref,breaklinks,colorlinks,citecolor=eccvblue]{hyperref}
\usepackage{orcidlink}

\begin{document}

\title{The Role of Generative Systems in Historical Photography Management:\\A Case Study on Catalan Archives} 

\titlerunning{Generative Systems in Historical Photography Management}

\author{
Èric Śanchez\orcidlink{0009-0000-7984-9856} \and Adrià Molina \thanks{Corresponding Author} \orcidlink{0000-0003-0167-8756}\and
Oriol Ramos Terrades \orcidlink{0000-0002-3333-8812} }
\institute{
Computer Science Department \\
Universitat Autònoma de Barcelona \and Computer Vision Center \\
\email{eric.sanchezlo@autonoma.cat \\ \{amolina, oriolrt\}@cvc.uab.cat} }

\authorrunning{E. Sánchez \and A. Molina and O. Ramos}

\maketitle

\begin{abstract}
The use of image analysis in automated photography management is an increasing trend in heritage institutions. Such tools alleviate the human cost associated with the manual and expensive annotation of new data sources while facilitating fast access to the citizenship through online indexes and search engines. However, available tagging and description tools are usually designed around modern photographs in English, neglecting historical corpora in minoritized languages, each of which exhibits intrinsic particularities. The primary objective of this research is to study the quantitative contribution of generative systems in the description of historical sources. This is done by contextualizing the task of captioning historical photographs from the Catalan archives as a case study. Our findings provide practitioners with tools and directions on transfer learning for captioning models based on visual adaptation and linguistic proximity.
\end{abstract}

\section{Introduction}
The task of image captioning aims to produce a coherent sentence that describes the relationships and context of the visual attributes of an image. Incorporating image captioning models in historical management could offer advantages in human-hours of manual annotations in recognizing the elements of the scene and, consequently, causing delays in public access and indexing of the content. 

In this work, we explore the use case of the management of multimedia heritage by the \textit{Xarxa d'Arxius Comarcals} (XAC, Catalan for Network of County Archives). In order to grant access to the citizenship with all the correctly tagged archives on their public access portal\footnote{XAC: \href{http://arxiusenlinia.cultura.gencat.cat}{arxiusenlinia.cultura.gencat.cat}}, extensive research and writing by the archivists is required. These publications can help find photographs of ancestors or poorly documented cultural celebrations that have not survived to the present day. Therefore, it is a duty to make these resources available to society as soon as possible and with the quality they require.

\begin{figure}[t]
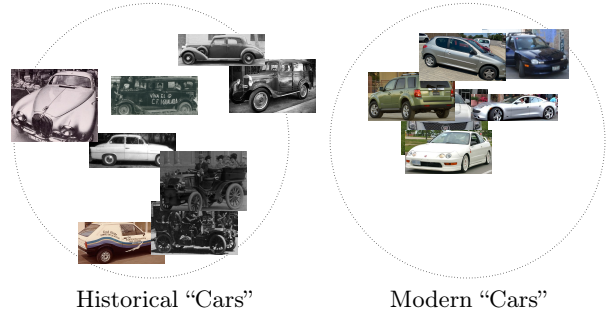

    \centering
    \begin{tabular}{cc}
         \resizebox{!}{0.3\textwidth}{\input{figures/cotxes/antics}}&
         \resizebox{!}{0.3\textwidth}{\input{figures/cotxes/moderns}}\\
         Historical ``Cars'' & Modern ``Cars''

    \end{tabular}
    \caption{Example where the visual attributes associated with the token ``car'' exhibit a higher visual variance in historical (left) than in modern (right) photographs. }
    \label{fig:cotxes}
\end{figure}

However, the current limitations on this and other deep learning tasks are the imbalance in the use of different languages in the models and datasets available on the internet and the consequent lack of resources in minoritized languages, thus hindering the development of high-performance tools in these languages. This gap is widened because of the historical nature of the images, which shifts the association between visual elements and their textual correspondences, as seen in Figure \ref{fig:cotxes}.

As a consequence of multimedia content exponentially growing in recent years, the data used to train image captioning models tends to be recent and only represent a biased portion of all existing photographs. As a consequence,such models perform worse on data from an earlier historical domain due to differences in elements such as vehicles, buildings and cities, and clothing, to name a few. 

Therefore, the objectives of this are to address the following research questions:
\begin{enumerate}
    \item \textbf{(RQ1)}: Are image generation models optimal for adapting captioning models to historical sources of data containing intrinsic demographic and historical properties?
    \item \textbf{(RQ2)}: Is text generation all you need for training models that can naively operate on minoritized languages?
    \item \textbf{(RQ3)}: Does language modeling fine-tuning operate proportionally to linguistic proximity? 
\end{enumerate} 

Thus, this work focuses on adapting image description generation models to specific historical and linguistic contexts. By using temporally adapted synthetic datasets and pretrained models in languages with more resources, we aim to improve the accuracy and relevance of descriptions of captioning systems. This research also utilizes a large amount of data from multiple languages to train the models, ensuring broader coverage. Ultimately, it ablates on the role of synthetic visual and text features in bridging the gap of historical and linguistically biased distributions in the context of the Catalan county archives.

\section{Related Work}
% Crec que aquí haurem de fer el major canvi, aqui no venem un model de captioning
% S'hauria de parlar coses com l'aina etc
\subsection{Image Captioning}
\textit{Image captioning} is a descriptive task that integrates computer vision and natural language processing to generate textual descriptions of images. The first deep learning models designed for the task of \textit{image captioning} \cite{Vinyals_2015_CVPR, Karpathy_2015_CVPR, Anderson_2018_CVPR} consisted of a CNN \cite{krizhevsky2012imagenet} to extract image features and the use of an RNN \cite{rumelhart1986learning} or LSTM \cite{hochreiter1997long} for text generation. Later, attention mechanisms and \textit{Transformers} \cite{Vaswani_Transformer_2017} emerged, providing additional efficiency in feature extraction from images and especially in text generation. A different perspective on the task of associating images with text is provided by \cite{gomez2017selfsupervisedlearningvisualfeatures} and, later on, CLIP \cite{radford2021learning}, which projects images and texts into a common vector space where similar pairs are closer than dissimilar ones, thereby allowing the calculation of a proximity value between them based on distance. This is useful for assessing the quality of model predictions and training data. 

One of the main challenges in working with historical images is temporal and stylistic variability. The \textit{Vision Transformer} (ViT) architecture \cite{dosovitskiy2021image} has proven effective in computer vision tasks, but adaptation to historical images requires additional techniques such as weight adaptation with specific data and data augmentation. Additionally, some authors suggest that the generation of synthetic datasets of historical images provides a robust basis for training models when real data is scarce \cite{Bartz_2022}. Capturing the intrinsic properties of historical photographs is a necessary condition for synthetic data generation. In this study, we will address the convenience of such approaches with respect to text-based generation. 

Additionally, recent trends on multi-modal large language modeling \cite{yin2024surveymultimodallargelanguage} showed an excelling performance in downstream tasks, including multi-lingual vision and text reasoning\cite{openai2024gpt4technicalreport}. However, this case of study will neglect the existence of LLMs, as they support an unreasonable economical and environmental burden that heritage and governmental institutions, such as XAC, cannot afford as a means for automation (see Section \ref{sec:synth}, where the additional cost of text processing in non-English languages due to sub-efficient language processing is empirically assessed).

In the multilingual context, \textit{Transformers} like mBERT (Multilingual BERT) \cite{devlin-etal-2019-bert} and XLM-R \cite{conneau2020unsupervised} have been widely used to generate descriptions in multiple languages. These models demonstrate that training in multiple languages helps improve performance in languages with fewer training samples. Moreover, adapting the weights for specific languages allows maintaining a smaller model size, as it does not need to capture information for all languages simultaneously, but can be pretrained on resource-rich languages and transfer the information to the resource-scarce language, achieving similar performance more efficiently \cite{cruz-2019-evaluating}. Data augmentation through automatic translation techniques has also been explored \cite{xia2019generalized} with performance improvements in resource-scarce languages. 

In conclusion, the combination of historical and linguistic domain adaptation is essential to improve the generation of descriptions for archival images, but literature lacks of success and failure realistic scenarios to quantitatively assess the degree of correctness of such generations in recreating the specific conditions of data tied to historical and sociological cues, as pointed out by \cite{stacchio2023analyzing}.

\subsection{The \textit{Xarxa d'Arxius Comarcals}}

As it has already been introduced, this work contextualizes image captioning as a case study for the integration of descriptive systems in automated historical photography management. The \textit{Xarxa d'Arxius Comarcals} is a cultural heritage institution that dates back to 1932 \cite{xacquisom}. The function of regional archives is the conservation, management, and dissemination of documentary heritage from its original environment, thus avoiding provincial concentrations.

Their first steps towards the integration of digital humanities in their strategy and development, started in 2005 \cite{expedients} by extracting information from francoist border files in an effort for automated recovery of victims during such historical periods. After that pioneering collaboration, the XAC reprised their collaboration, now in  a regular basis, of extracting information from historical census records \cite{xarxes}, aiming to reconstruct and model the relationships and demographic behaviors of our ancestors. Other projects involved extractive tasks, such as named entity recognition, on marital census \cite{i2021fons, romero2013esposalles}, serving the same purpose. 

In previous applications, the specificity of the data and the use of fixed-structure sentences served as a proxy to facilitate information extraction (see \cite{romero2013esposalles}). In this work, we explore the feasibility of XAC transitioning from extractive to descriptive tasks in digital humanities applications. This shift involves addressing challenges associated with an open-domain Catalan vocabulary, for which descriptive systems are not widely available, and working with photographs ranging from the 19th to the early 21st century.

\section{Architecture}
\label{sec:arch}
In this section, we introduce the technical details of the implementation on which we will conduct our study. Since LLMs are not well suited for heritage and government institutions, both from an economical and environmental perspective, we seek compact models that can fit on a reasonable scale. 

\begin{figure}[h]
    \centering
    \includegraphics[width=0.8\linewidth]{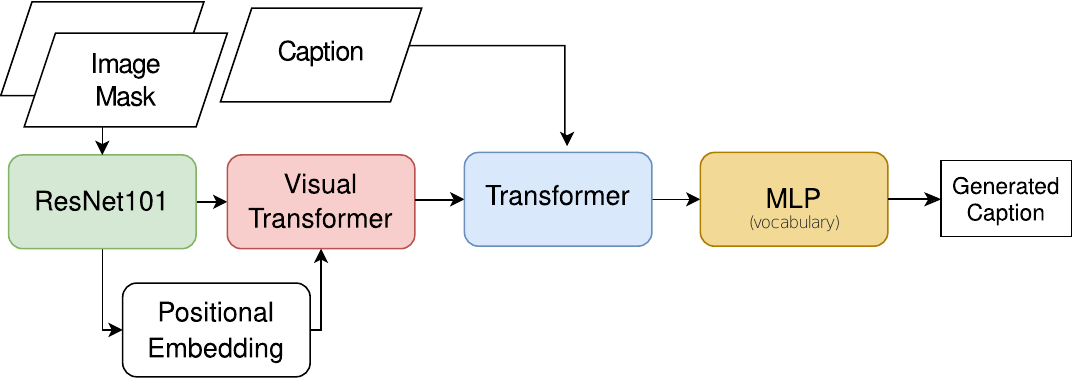}
    \caption{CATR model architecture}
    \label{fig:model-CATR}
\end{figure}

The \textit{Image Captioning} model used as the basis for the project is called CATR (CAption TRansformer)\footnote{CATR GitHub repository: \href{https://github.com/saahiluppal/catr}{https://github.com/saahiluppal/catr}}. It consists of a Convolutional Neural Netowork (CNN) followed by a \textit{vision transformer} block in the encoder and a textual \textit{Transformer} in the decoder part. The main difference between the encoder part of this model and the Visual Transformer\cite{dosovitskiy2021image} is the way it obtains the embedding of the input image.

First, instead of splitting the image into patches, the image is processed by a \textbf{ResNet}\cite{He_2016_CVPR} and the output features are used along with their positional embedding. The visual features are contextualized through an encoder (\textbf{visual transformer}) and, secondly, this feature conditions an autoregressive \textbf{transformer decoder}. This decoder produces tokens given the previous state and conditioned to a set of visual features. The generated tokens are projected into the vocabulary size through a multi-layered perceptron (\textbf{MLP}).

\section{Datasets}
%\section{Experimental Setup}
In this section, we introduce the experimental setup that this study will follow to quantitatively assess the contribution of both text and image generative systems in adapting captioning models to historical and Catalan data from XAC. In Section \ref{sec:data}, we will introduce and analyze the different datasets used to train and fine-tune our models, including the set of XAC images for this case study and the generation of synthetic datasets that will support this research endeavor. Lastly, in Section \ref{sec:qualitative}, we provide a brief qualitative image assessment on the synthetically generated images.

\subsection{Benchmark Datasets}
% Aquí introduir els datasets que es faran servir
In this study, we leverage several datasets serving different purposes. First, we consider COCO\cite{lin2015microsoft} as one of the best established research datasets for modern image captioning. COCO will serve as the main comparative mechanism and baseline, as its size and data quality are on a sweet spot for research and reproducibility in every laboratory and institution. Secondly, we leverage the Ducth language from CrossModal3600  \cite{thapliyal_crossmodal-3600_2022}.

In our case study, $\sim$30K XAC images have been yielded to our research as a preliminary extraction for studying the feasibility of incorporating automated description systems into their production scheme.

\begin{figure}[t]
    \centering
    \begin{tabular}{cccc}
         \includegraphics[width=0.21\textwidth]{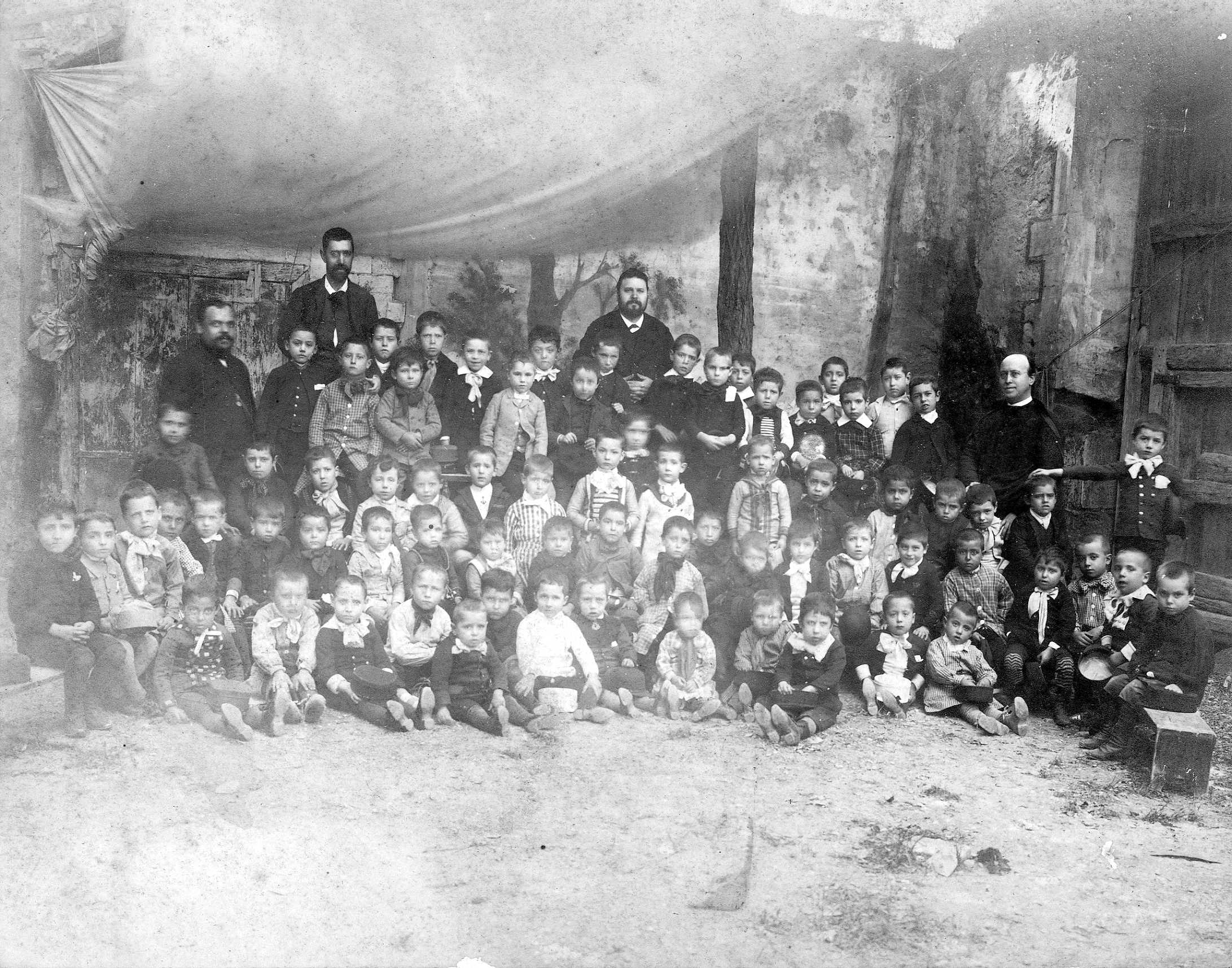}&
         \includegraphics[width=0.21\textwidth]{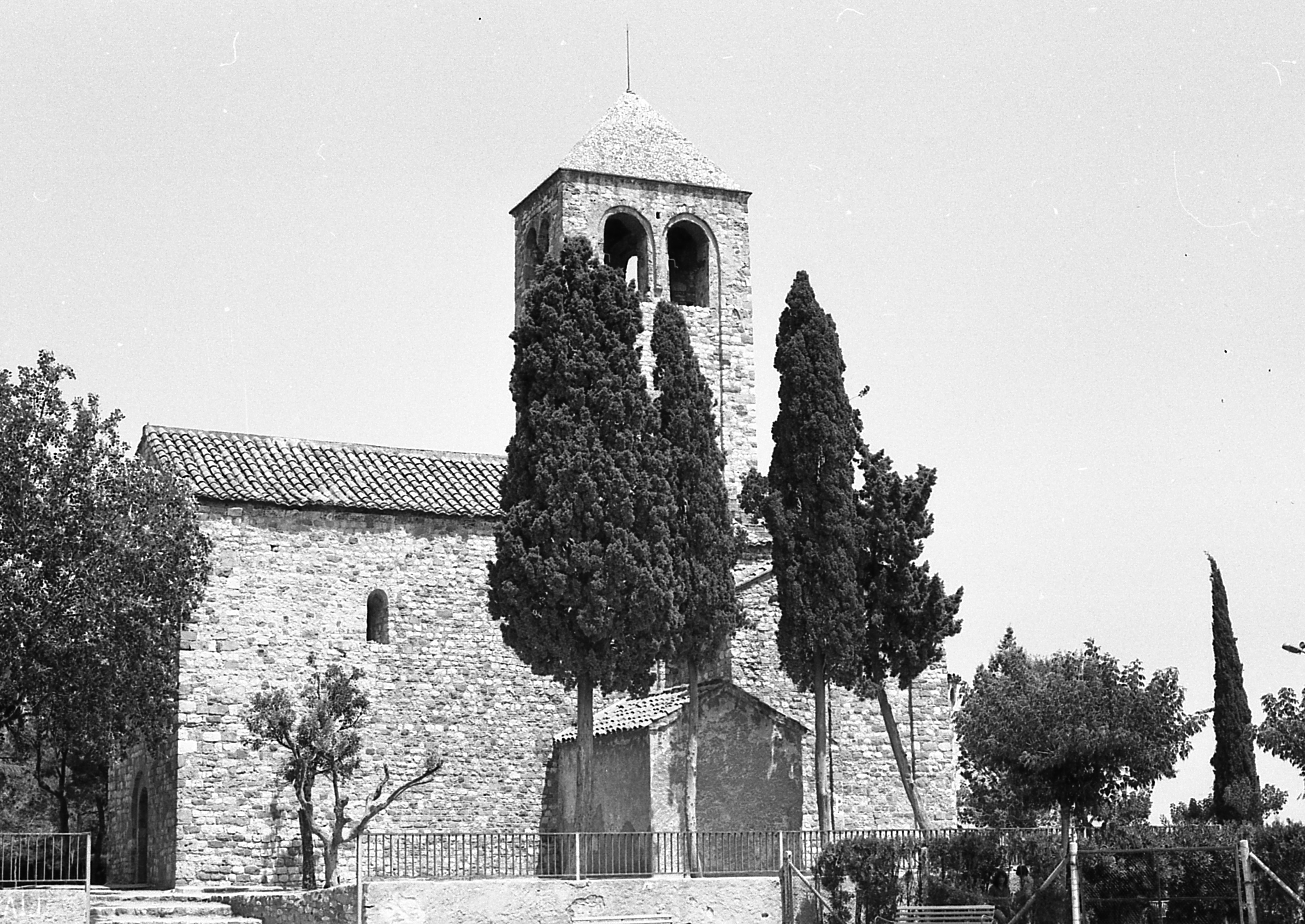} &\includegraphics[width=0.21\textwidth]{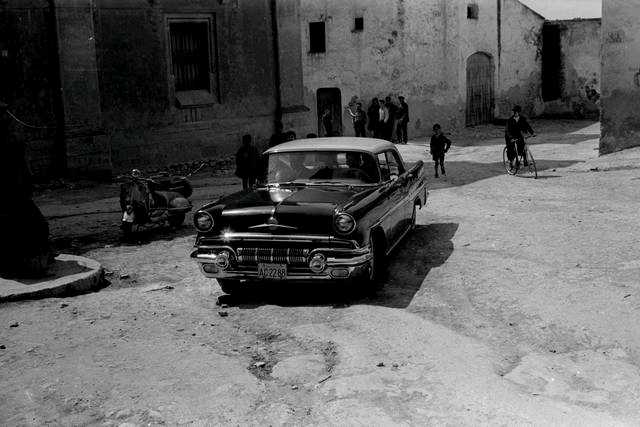} &\includegraphics[width=0.21\textwidth]{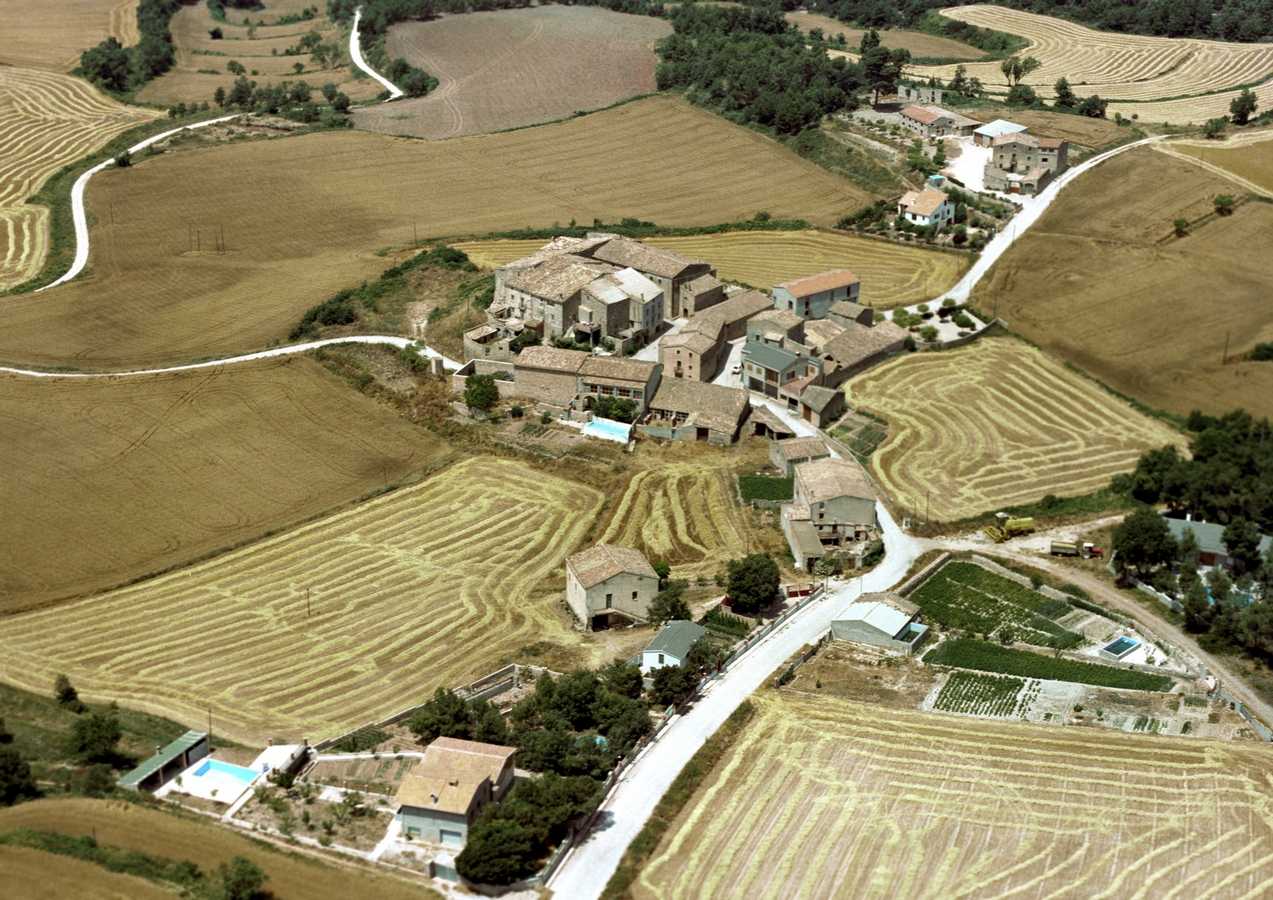}  \\
\begin{minipage}[t]{0.23\textwidth}
    \input{figures/xac_samples/ACAP_1022712_021101730000255}
\end{minipage}&
\begin{minipage}[t]{0.23\textwidth}
    \input{figures/xac_samples/ACVOC_1305082_090036530001232}
\end{minipage}&
\begin{minipage}[t]{0.23\textwidth}
    \input{figures/xac_samples/AHT_397557_1499,0155}
\end{minipage}&
\begin{minipage}[t]{0.23\textwidth}
    \input{figures/xac_samples/AHT_647816_7,0001}
\end{minipage}

    \end{tabular}
    \caption{Samples from the XAC collection, we observe an important presence of named entities. The collection also shows high temporal and content diversity.}
    \label{fig:xac_examples}
\end{figure}

As it can be seen in Figure~\ref{fig:xac_examples} the XAC collection contains captions with a high presence of named entities (names, locations, etc.). This is due to the historical relevance of the institution in categorizing and preserving heritage at the finest granularity. Because named entities are usually arbitrary pieces of information rarely associated with recognizable patterns in images, this is avoided in curated research datasets such as COCO.

The incorporation of named entities in captioning systems is a widely explored field \cite{biten_good_2019,nguyen2023show,zhao2021boosting} that usually requires external knowledge to be solved. Therefore, in the XAC dataset we apply the pre-processing step of named entity masking using the Catalan transformer model from \texttt{SpaCy} \footnote{\url{spacy.io/models/ca}}. The detected named entities are then masked with  \texttt{[PER]}, \texttt{[LOC]}, \texttt{[ORG]} and \texttt{[MISC]} regarding their detected labels (person, location, organization, and miscellany).

Additionally, the XAC dataset contains historical and Catalan data, which promotes important performance gaps when fine-tuning from the previously introduced datasets containing modern images with, non-Catalan captions. Therefore, in this work we aim to explore the quantitative contribution of both image and text generation models in bridging this gap.

\subsection{Synthetically Generated Dataset}
\label{sec:synth}
Synthetic data is created with the aim of having a set of images halfway between the COCO and XAC datasets in terms of historical domain, that is, images artificially adapted to an older temporal domain. To generate these images, the Stable Diffusion XL model \cite{podell2023sdxl}, a text-to-image generation model, was used, taking the COCO descriptions as a reference. The strategy followed to obtain these images in a different historical domain was to use each of the COCO descriptions (one description per image, not all descriptions) as input to the model with an addition of the year we want the image to represent, specifically: \texttt{<caption> in <year>}.

The selected years are, as a controlled set, the same as those present in the Date Estimation in the Wild dataset \cite{Muller_DEW_2017}, from 1930 to 1999. The years are randomly chosen following a uniform distribution $U \sim (1930, 1999)$. This will help us explore how fine-tuned models behave on unseen years during training with respect to this set.

\begin{figure}[t]
    \centering
    \begin{tabular}{cc}
         \includegraphics[width=0.45\textwidth]{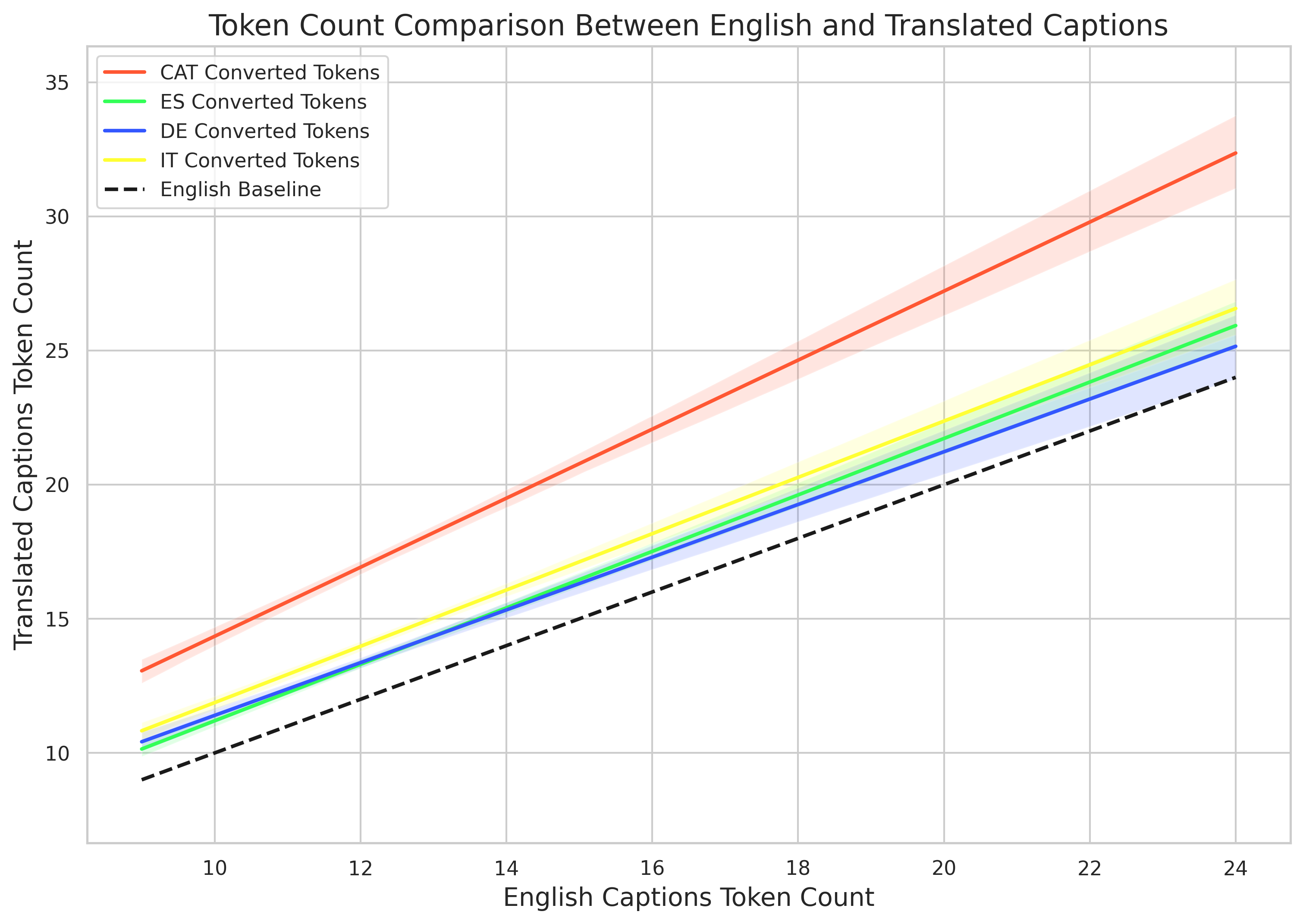}&\includegraphics[width=0.45\textwidth]{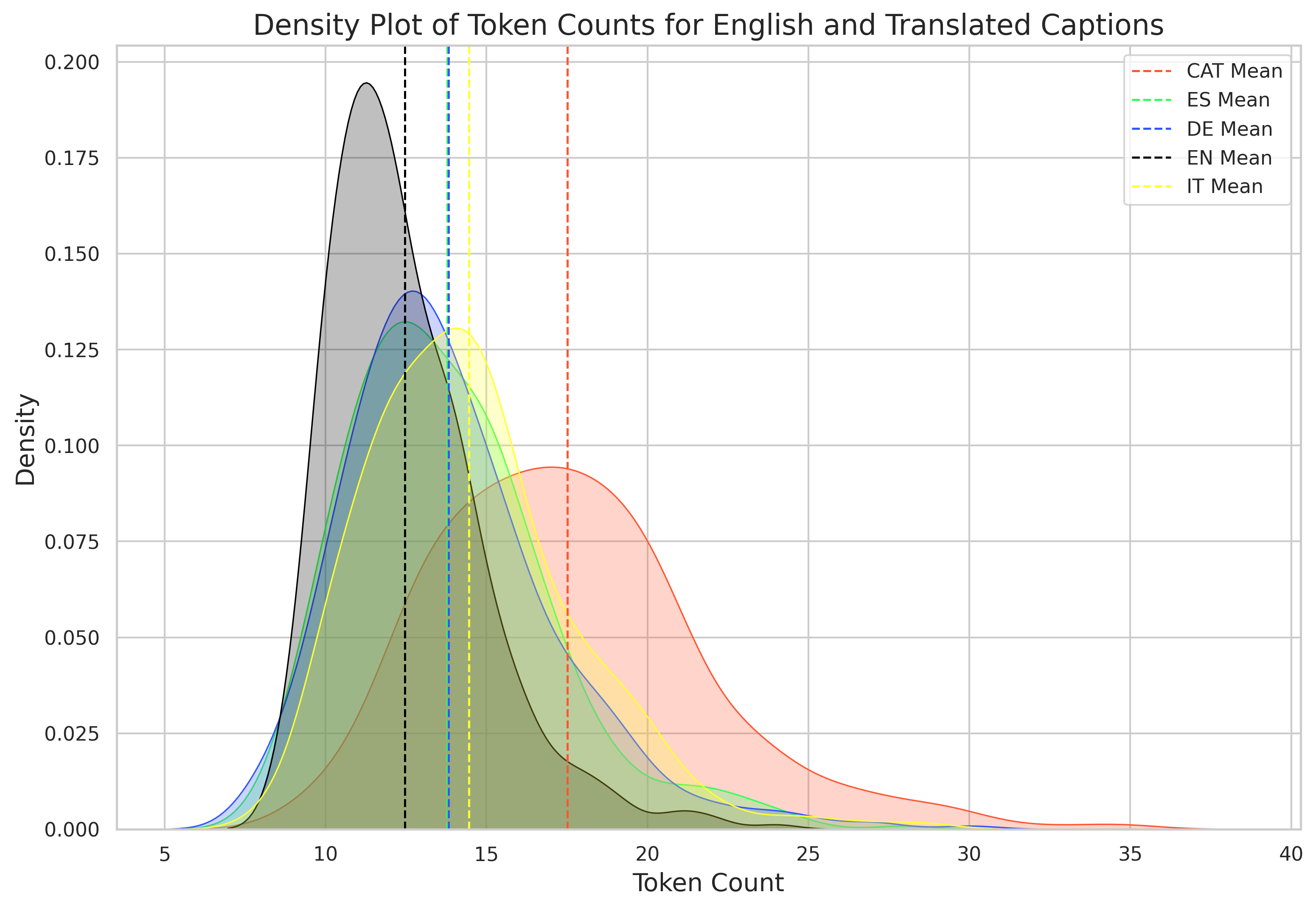}  
    \end{tabular}
    \caption{Gemma-2\cite{gemma_2024} tokenization count for English versus translated captions (left) and distribution of number of tokens per language (right). }
    \label{fig:num_tokens}
\end{figure}

On the other hand, in addressing text generations, COCO captions have been automatically translated to Catalan, Spanish, Italian, German, and Dutch from the descriptions using two different translation models. For Catalan, the Softcatalà translation model from English to Catalan \cite{huggingfaceSoftcatalaopennmtengcatHugging}, based on OpenNMT~\cite{klein-etal-2017-opennmt}, was used. For the other languages, the \texttt{M2M100\_418M} model \cite{fan2020englishcentric} was used, a translation model for 100 different languages that performs translation directly between the 9900 different language pair combinations without having to go through English. In the following sections, we will thoroughly explore the impact of text generation in improving our case study fine-tuning process. As it was previously mentioned, Figure~\ref{fig:num_tokens} suggests how LLMs tend to provide sub-efficient tokenization in any of the sampled lower resource languages, which fatherly motivates this research as a means to alleviate the unnecessary computational and economical burden of massively accessing an API of a non-native or generalist language model.

\subsection{Image Generation Qualitative Assessment}
\label{sec:qualitative}

\begin{figure}[!h]
    \centering
    \begin{minipage}{.45\linewidth}
        \centering
        \includegraphics[width=.9\linewidth]{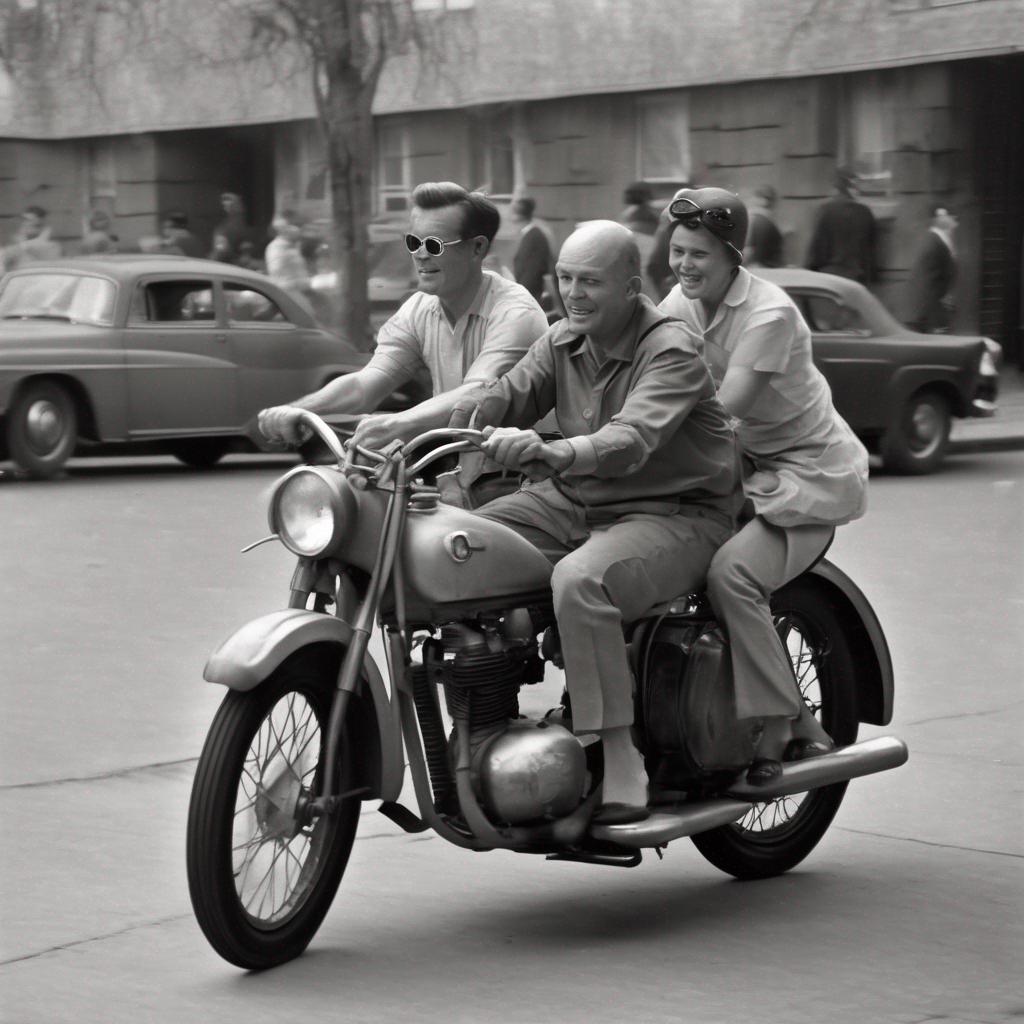}
        \caption{Poor representation of interactions.}
        \label{fig:hist_sd0}
    \end{minipage}%
    \hfill
    \begin{minipage}{.45\textwidth}
        \centering
        \includegraphics[width=.9\linewidth]{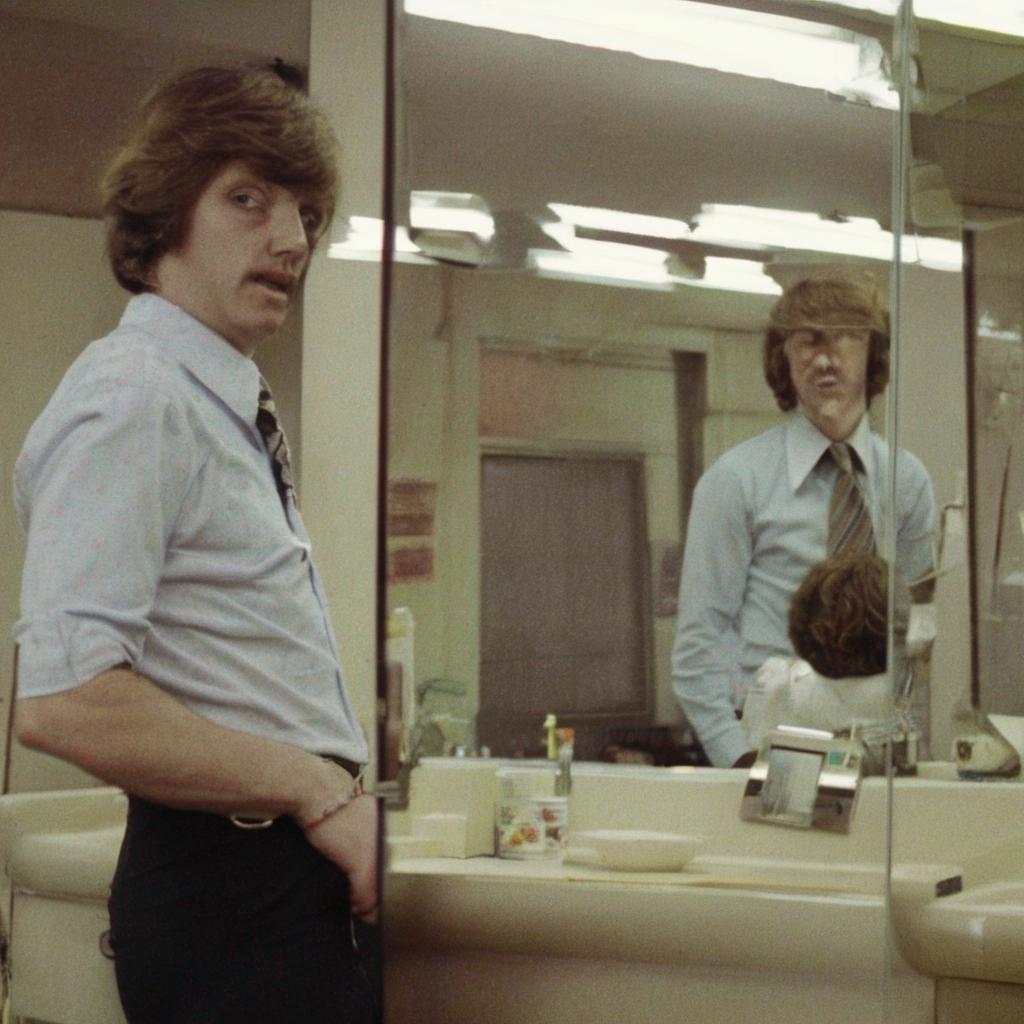}
        \caption{Poor representation of faces.}
        \label{fig:hist_sd1}
    \end{minipage}
\end{figure}

The Stable Diffusion XL model \cite{podell2023sdxl} is trained on the LAION 5B dataset \cite{LAION5B2022} and has certain biases from both the data it was trained on and the model itself. The model has difficulties in representing faces and limbs of people, and sometimes some objects interact strangely with each other (see Figures \ref{fig:hist_sd0} and \ref{fig:hist_sd1}). The images generated with this noise complicate the task of recognition and subsequent description by the image description generation model.

To identify these biases, a test was carried out. The goal is to identify the objects or attributes that are most often incorrectly labeled and to conduct a qualitative study on whether this is due to biased image generation or a deficiency in the image description generation model.

\begin{figure}[t]
    \centering
    \includegraphics[width=.9\linewidth]{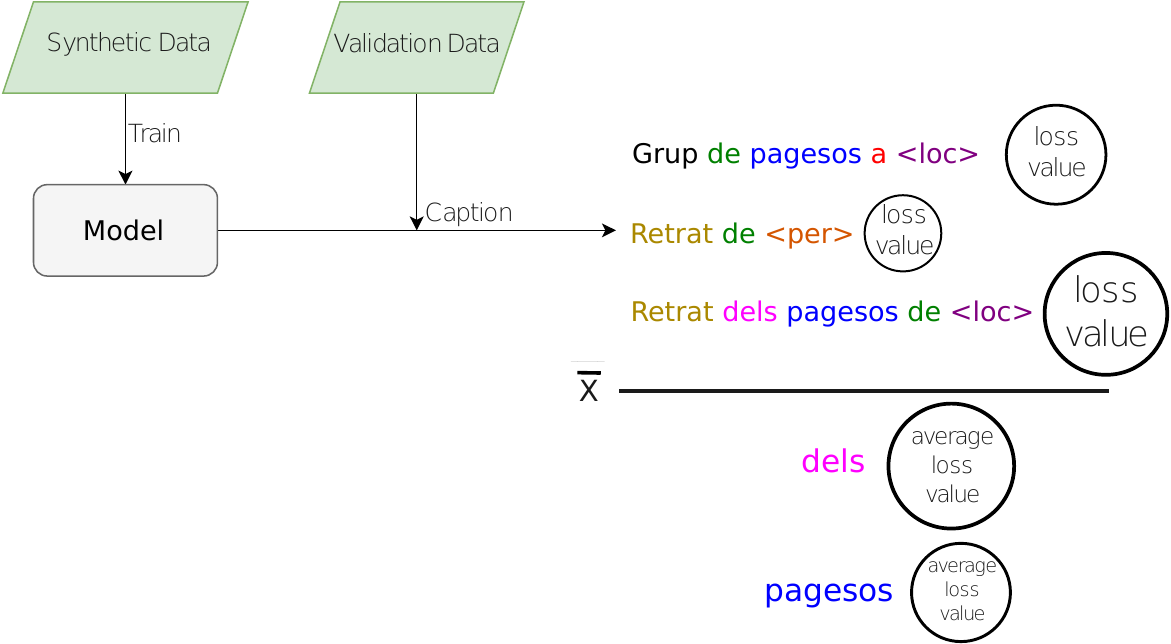}
    \caption{Procedure for obtaining the terms with the highest average loss function value.}
    \label{fig:qualitative-study}
\end{figure}

A model was trained using the generated images and the associated original COCO descriptions, which was then used to generate predictions for the descriptions of the validation partition. For each generated description, the loss function relative to the actual description and all the tokens appearing in it were recorded. Finally, for each token, the average loss function values of all samples where it appears were calculated (see \figureautorefname~\ref{fig:qualitative-study}). Below are the 28 words with the highest average loss function values. \\

%\texttt
\begin{minipage}{\textwidth}
    \centering{\tt  \{suggests, immediate, personi, closest, similarly, benefit, Barclay, institution,\\ thirty, operates, Lower, centered, Outside, Cuba, whom, coins, higher, protein, Away,\\ victim, trouble, headquartered, danger, assault, bordered, writings, services, Silk\}}\\

\end{minipage}

\noindent From the above list, several semantic relationships can be observed in certain words:

\begin{itemize}
    \item \textbf{Compositional relationships}: the words \textit{immediate, closest, Lower, centered, Outside, higher} express compositional qualities of the entities represented in the original image. These relationships are lost in the image generation process, and consequently, the description generation model does not observe these relationships.
    \item \textbf{Abstract concepts}: the terms \textit{suggests, similarly, benefit, Cuba, victim, services} deal with concepts that are not explicitly representable but must be inferred with additional context or by interpreting more complex relationships between the entities that appear. This interpretation is difficult for the generator model to perceive and subsequently represent, so this information is lost.
    \item \textbf{Brands and materials}: the words \textit{Barclay, Silk} may not be well represented in the generated image for various reasons; in the case of the brand, due to the diffusion model's lack of knowledge of its existence or the difficulty in generating a logo seen in few training samples. On the other hand, materials like silk require high image quality to be clearly distinguishable, which is not present in either the generated image or the processed image received by the generator model.
\end{itemize}

Lastly, in Table \ref{table:datasets}, we conducted a brief meta-analysis where we show how the XAC data exhibits the lowest \texttt{CLIP-Score}, which stands to the average distance of the image-caption pairs of each dataset using CLIP embeddings\cite{radford2021learning}. Although it serves as a sign of the high quality of the captions, it goes in detriment of diversity, expressed as the standard deviation of the distances, \texttt{CLIP-std}.

\begin{table}[b]
\centering
\begin{tabular}{|l|l|l|l|l|l|l|}
\hline
\textbf{Dataset}      & \textbf{Size} & \textbf{Domain} & \textbf{Languages} & \textbf{CLIP-score} & \textbf{CLIP-std} \\ \hline
MS COCO               & 591k  & $\dagger$  & en & 0.121         & 3.688         \\
Crossmodal3600        & 261k  & $\dagger$  & 36 & 0.092         & 3.790         \\
Synthetic dataset  & 591k  & 1930-1999* & 5  & 0.137         & 3.688         \\ 
\textbf{XAC}  (ours)                 & 29k   & 1880-2020  & ca & 0.073 (0.063) & 3.480 (3.466) \\ \hline
\end{tabular}
\caption{Meta-analysis of the datasets. (·) Variants with masked \textit{named entities}, * synthetic data, $\dagger$ information not availible.}
\label{table:datasets}
\end{table}

\section{Experiments and Results}
In this section we perform a deep discussion on the results of our case study. Here, we pose asses the initial research questions regarding image (\textbf{RQ1}) and text (\textbf{RQ2}) generative systems contribution to domain adaptation and sensitivity of language models to cognitive lexical rules (\textbf{RQ3}).

In Section \ref{sec:hist_domain} we discuss the interplay between \textbf{RQ1} and \textbf{RQ2}, on which we draw the line through vision and language through an ablation study on generated pretraining data for both modalities. In Section \ref{sec:lang_domain}, we propose a comparison of transferring knowledge from source and target languages via language clusters based on lexical rule matching as a means to address \textbf{RQ3}.
\subsection{Experimental setup}
%\subsection{Technical Details}
\label{sec:tech}
As it was previously introduced during Section \ref{sec:arch}, this work aims to explore its hypothesis within the context of reasonable-scale descriptive visual systems. Therefore, all experiments, training, and inference are conducted on single \texttt{NVIDIA Titan Xp} GPUs, \texttt{CUDA Version: 11.2}, with a maximum training time of 4 days.

\begin{table}[t]
    \centering
    \begin{tabular}{|l|c|}
        \hline
        \textbf{Definition} & \textbf{Value} \\ \hline
        Dimension of the hidden layer of the MLPs & $512$ \\
        Maximum size of the feature vector  & $256$ \\
        Height and width of the feature maps & $19px$ \\
        Dimension of the hidden layer of the visual \textit{transformer} & $256$ \\
        Maximum size of the \textit{embedding} & $128$ \\
        Number of attention heads & $8$ \\
        Number of \textit{encoder} and \textit{decoder} layers & $6$ \\
        Vocabulary size & $119769$ \\
        Number of feature channels & $2048$ \\
        Height and width of the input image and mask to the model & $299px$ \\ \hline
    \end{tabular}
    \caption{Empirically defined model hyperparameter values.}
    \label{tab:variables-model}
\end{table}

The defined hyperparameters of the model can be found at \ref{tab:variables-model}, where a brief description of each variable and its selected value is defined. 
\label{sec:data}

%\section{Experiments}

% Aqui podem parlar de que l'error acomulat ve de coses mal generades (lo de que certes paraules mostraven pèrdues més grans)
\subsection{Historical Domain Adaptation}
\label{sec:hist_domain}
\begin{table}[b]
    \centering
    \begin{tabular}{|cc|cc|}
    \hline
    \multicolumn{2}{|c|}{Pretraining on COCO} & \multicolumn{2}{c|}{Finetune on XAC} \\
    \textbf{Synthetic Images} & \textbf{Synthetic Captions} & \textbf{CIDEr} $\uparrow$ & \textbf{BLEU4} $\uparrow$ \\ \hline
    \xmark           & \xmark             & 0.497            & 0.133            \\ 
    \xmark           & \cmark             & \textbf{0.697}   & \textbf{0.151}   \\ 
    \cmark           & \xmark             & 0.559            & 0.140            \\ 
    \cmark           & \cmark             & 0.600            & 0.141            \\ \hline
    \end{tabular}
    \caption{Ablation study on the performance of fine-tuned historical captioning models when using synthetic images and captions as pre-training data.}
    \label{table:ablation-synthetic-ner}
\end{table}

Upon inspection of Table \ref{table:ablation-synthetic-ner}, several notable trends emerge. Firstly, the use of synthetic captions proves to be a dominant strategy in knowledge transfer during the fine-tuning step, unlike the case with synthetic images. Secondly, while the use of synthetic and historically adapted images shows advantages over modern photographs even without automatic caption translation, the accumulation of noise from both synthetic approaches leads to a decrease in CIDEr and BLEu4 scores compared to the top-performing strategy: using modern images with synthetic captions (\textbf{RQ2}). This behavior suggests that much of the error in historical captioning systems lies within the language modeling rather than the visual encoder, which effectively generalizes its features from modern photographs. Nevertheless, the initial hypothesis that historical variance poses a challenge remains valid, as synthetically generated images introduce some positive bias when used independently as a pre-training stage.

\begin{figure}[h]
    \centering
    \begin{minipage}{.45\linewidth}
    
        \includegraphics[width=\textwidth]{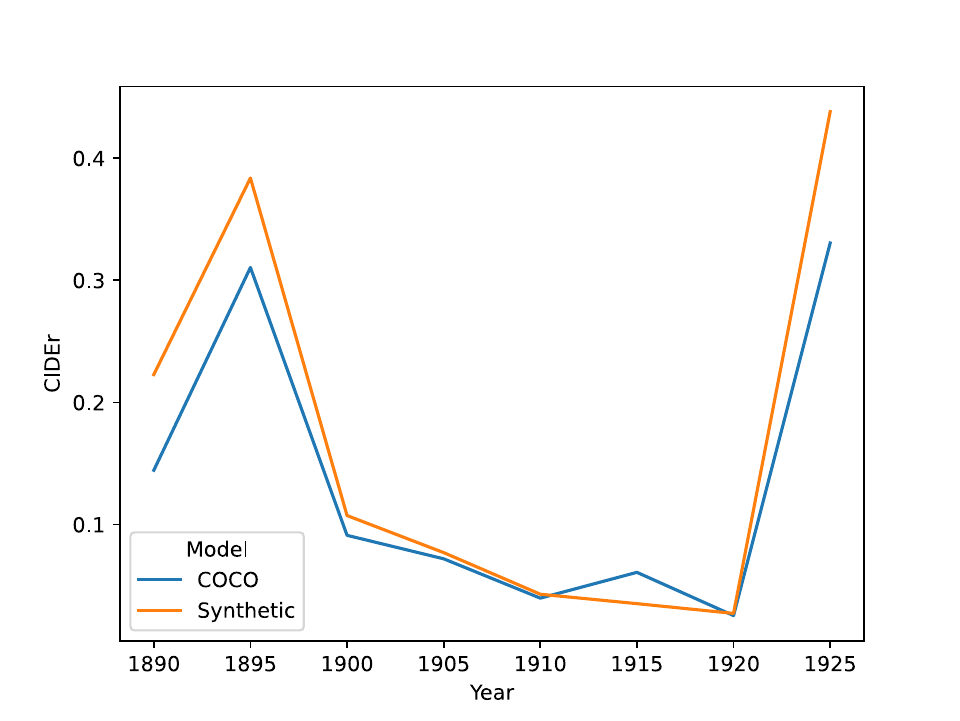}%
        \caption{CIDEr as a function of year for data prior to 1930, evaluated on the pretrained model using synthetic images and descriptions.}
        \label{fig:to-1930-cider}
    \end{minipage}%
    \hfill
    \begin{minipage}{.45\linewidth}
        \includegraphics[width=\linewidth]{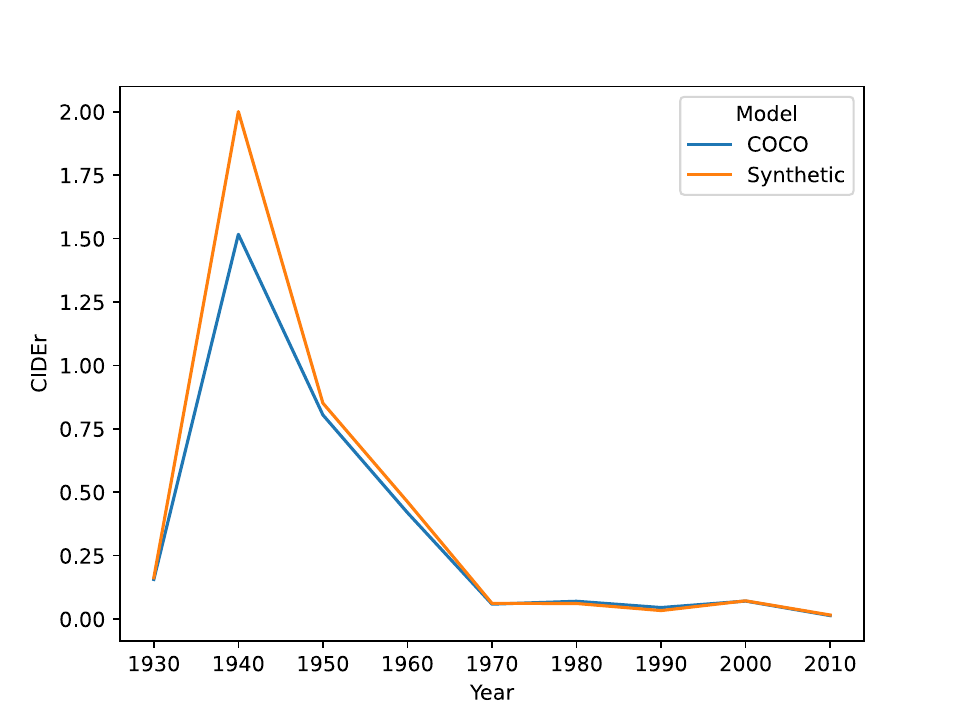}%
        \caption{CIDEr as a function of year for data after 1930, evaluated on the pretrained model using synthetic images and descriptions.}
        \label{fig:from-1930-cider}
    \end{minipage}%    
\end{figure}

Moreover, as observed in \figureautorefname~\ref{fig:to-1930-cider}, the model incorporating synthetic images tends to slightly outperform on images outside the temporal range covered by the synthetic dataset (1930-1980). Conversely, as shown in \figureautorefname~\ref{fig:from-1930-cider}, the synthetic model stabilizes its performance on dates within the training temporal range. This performance convergence on later years suggests that while the use of synthetic images may aid in transferring knowledge in specific scenarios, its effectiveness heavily relies on temporal biases already present in the model rather than capturing a realistic representation of content relevant to each historical period. This approach assumes uniformity in captions across history; however, as observed by \cite{salem2016analyzing}, content depicted in different time periods may not exhibit such uniformity. In conclusion, image generation appears unable to maintain the necessary historical coherence to consistently prove effective in historical domain adaptation (\textbf{RQ1}), but might incorporate colors, textures, and some bias that promotes its average success (see Table \ref{table:ablation-synthetic-ner}).

\begin{figure}[b]
    \centering
    \includegraphics[width=0.5\linewidth]{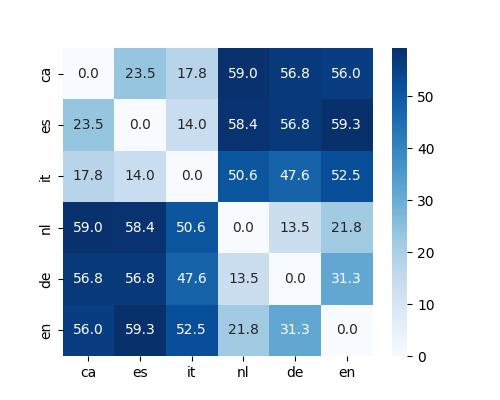}
    \caption{Language proximity extracted from the eLinguistics project\protect\footnotemark.}
    \label{fig:lang-proximity}
\end{figure}
\footnotetext{eLinguistics blogpost: \href{http://www.elinguistics.net/Language_Timelines.html}{www.elinguistics.net/Language\_Timelines.html}}

\subsection{The Role of Language}
\label{sec:lang_domain}

\begin{table}[h]
\centering
\begin{tabular}{|c|c|c|c|}
\hline
\textbf{Source language} & \textbf{Target language} & \textbf{CIDEr} $\uparrow$ & \textbf{BLEU4} $\uparrow$ \\ \hline
-               & ca              & 2.196            & 0.276           \\ 
es              & ca              & \textbf{2.411}   & \textbf{0.292}   \\ 
de              & ca              & 2.367            & 0.283            \\ 
es, it          & ca              & 2.236            & 0.274            \\ 
en, de*         & ca              & 2.308            & 0.282            \\ \hline
-               & nl              & 0.245            & 0.025            \\ 
es              & nl              & 0.248            & 0.025            \\ 
de              & nl              & 0.252            & \textbf{0.026}   \\ 
es, it          & nl              & 0.251            & \textbf{0.026}   \\ 
en, de*         & nl              & \textbf{0.261}   & \textbf{0.026}   \\ \hline
\end{tabular}
\caption{Results on fine-tuning on a "target language" after pre-training on a "source language". The symbol (-) stands for no pre-training. XAC data test split is used as validation for Catalan (cat), and Crossmodal3600 for Dutch (nl). *Deutsch (de) and English (en) data are heavily biased towards English in quantity.}
\label{table:pre-training}
\end{table}

Following \cite{elinguistics}, we consider in Figure~\ref{fig:lang-proximity} two clusters of languages. First, we synthetically generate captions in Spanish (es) and Italian (it) as higher resource languages that should alleviate the scarcity of historical Catalan (cat) sources. In order to contrast this hypothesis of language modeling operating proportionally to lingüistic proximity, we pose a similar challenge with the Ducth (nl) language, which is clustered to English and German (de). 

Consequently, in Table \ref{table:pre-training}, we examine the role of using different pre-training languages to adapt to both Catalan and Dutch. At first glance, the conclusions drawn from the experiment seem reasonable: Spanish is the best pre-training language for Catalan, while the English-German cluster proves more convenient for Ducth. However, a deeper inspection demonstrates that the scale is playing a crucial role in this scenario. First, using German as pre-training language for Catalan slightly outperforms the Spanish-Italian cluster. However, using the English-German cluster for Catalan archives halves the final performance. Therefore, addressing \textbf{RQ3}, language modeling is, as expected, subjected to linguistic rules, but if the process of pre-training conveys in contradictory gradients (using both languages on a similar scale), the final performance slightly decreases. In the case of the English-German cluster outperforming in Dutch finetuning, we can attribute it to the dominance of English data in the pre-training stage. In such a scenario, the quantity of data compensates for the contradiction of the gradients during training.

\section{Conclusions and Further Work}

This work deepens into the role of image and text generation in adapting descriptive systems to unseen historical and linguistic domains. Framed within the context of Catalan historical archives, this research highlights several key findings:

\begin{itemize}
    \item \textbf{Bias in Pretraining:} Although some positive bias is introduced during the pretraining stage using image generation systems, it proves more advantageous to use natural images with translated captions. This is likely due to the accumulated noise from both generative approaches during the training stage.
    \item \textbf{Historical Context Sensitivity:} Current image generation models are not capable of introducing fine-grained historical cues. The dependence of historical photographs on contextual attributes exacerbates the gap between text and image generation, leading to less effective captioning.
    \item \textbf{Language Proximity:} Image captioning models are quantitatively sensitive to language proximity. These models show higher performance when pretrained on similar languages, even when synthetic captions are used, indicating the importance of linguistic similarity in pretraining datasets.
\end{itemize}

In summary, the combination of natural images and translated captions, attention to historical context, and leveraging language proximity are critical for improving the performance of descriptive systems in historical and linguistically diverse archives.

However, this work is limited by the reliance of generative systems in order to maximize the performance of the image captioning task. This leads to bias, which ties the final performance to the ability of the generative system to capture reality. We highlight the necessity of exploring further domain adaptation techniques (such as multi-task learning) as model-independent alternatives.

Additionally, measuring the impact of such models in the context of Catalan archives is crucial to proving that all the efforts pursued by the digital humanities field are worthwhile. This could be extended to other minoritized languages in a wider study, which would require the coordinated effort of a significant number of heritage institutions and technological centers.

\section*{Acknowledgements}
This work has been partially supported by the Spanish project PID2021-126808OB-I00, Ministerio de Ciencia e Innovación, the Departament de Cultura of the Generalitat de Catalunya, and the CERCA Program / Generalitat de Catalunya. Adrià Molina is funded with the PRE2022-101575 grant provided by MCIN / AEI / 10.13039 / 501100011033 and by the European Social Fund (FSE+).

\tiny{
\bibliographystyle{splncs04}
\bibliography{main}}

\begin{thebibliography}{10}
\providecommand{\url}[1]{\texttt{#1}}
\providecommand{\urlprefix}{URL }
\providecommand{\doi}[1]{https://doi.org/#1}

\bibitem{Anderson_2018_CVPR}
Anderson, P., He, X., Buehler, C., Teney, D., Johnson, M., Gould, S., Zhang, L.: Bottom-up and top-down attention for image captioning and visual question answering. In: Proceedings of the IEEE Conference on Computer Vision and Pattern Recognition (CVPR) (June 2018)

\bibitem{Bartz_2022}
Bartz, C., Raetz, H., Otholt, J., Meinel, C., Yang, H.: Synthesis in style: Semantic segmentation of historical documents using synthetic data. In: 2022 26th International Conference on Pattern Recognition (ICPR). IEEE (Aug 2022). \doi{10.1109/icpr56361.2022.9956471}, \url{http://dx.doi.org/10.1109/ICPR56361.2022.9956471}

\bibitem{biten_good_2019}
Biten, A.F., Gomez, L., Rusiñol, M., Karatzas, D.: Good {News}, {Everyone}! {Context} driven entity-aware captioning for news images (Apr 2019), \url{http://arxiv.org/abs/1904.01475}, arXiv:1904.01475 [cs]

\bibitem{conneau2020unsupervised}
Conneau, A., Khandelwal, K., Goyal, N., Chaudhary, V., Wenzek, G., Guzmán, F., Grave, E., Ott, M., Zettlemoyer, L., Stoyanov, V.: Unsupervised cross-lingual representation learning at scale (2020)

\bibitem{expedients}
{Consell Comarcal de l'Alt Empordà}: L'arxiu comarcal col·labora amb un projecte del centre català de visió per computador. \url{https://www.altemporda.org/portal/component/content/article/189-noticies/3543-l-arxiu-comarcal-col-labora-amb-un-projecte-del-centre-catala-de-visio-per-computador}, accessed: 2024-07-20

\bibitem{cruz-2019-evaluating}
Cruz, J.C.B.B., Cheng, C.: Evaluating language model finetuning techniques for low-resource languages  (2019). \doi{10.13140/RG.2.2.23028.40322}, \url{http://rgdoi.net/10.13140/RG.2.2.23028.40322}

\bibitem{devlin-etal-2019-bert}
Devlin, J., Chang, M.W., Lee, K., Toutanova, K.: {BERT}: Pre-training of deep bidirectional transformers for language understanding. In: Burstein, J., Doran, C., Solorio, T. (eds.) Proceedings of the 2019 Conference of the North {A}merican Chapter of the Association for Computational Linguistics: Human Language Technologies, Volume 1 (Long and Short Papers). pp. 4171--4186. Association for Computational Linguistics, Minneapolis, Minnesota (Jun 2019). \doi{10.18653/v1/N19-1423}, \url{https://aclanthology.org/N19-1423}

\bibitem{dosovitskiy2021image}
Dosovitskiy, A., Beyer, L., Kolesnikov, A., Weissenborn, D., Zhai, X., Unterthiner, T., Dehghani, M., Minderer, M., Heigold, G., Gelly, S., Uszkoreit, J., Houlsby, N.: An image is worth 16x16 words: Transformers for image recognition at scale (2021)

\bibitem{xarxes}
{El Baix}: Projecte xarxes. \url{https://www.elbaix.cat/2017/06/07/larxiu-comarcal-presenta-el-projecte-xarxes-la-creacio-duna-xarxa-social-historica-del-baix-llobregat-i-lalt-penedes-en-el-marc-de-celebracio-del-dia-dels-arxius/}, accessed: 2024-07-20

\bibitem{openai2024gpt4technicalreport}
et.al, O.: Gpt-4 technical report (2024), \url{https://arxiv.org/abs/2303.08774}

\bibitem{fan2020englishcentric}
Fan, A., Bhosale, S., Schwenk, H., Ma, Z., El-Kishky, A., Goyal, S., Baines, M., Celebi, O., Wenzek, G., Chaudhary, V., Goyal, N., Birch, T., Liptchinsky, V., Edunov, S., Grave, E., Auli, M., Joulin, A.: Beyond english-centric multilingual machine translation (2020)

\bibitem{gomez2017selfsupervisedlearningvisualfeatures}
Gomez, L., Patel, Y., Rusiñol, M., Karatzas, D., Jawahar, C.V.: Self-supervised learning of visual features through embedding images into text topic spaces (2017), \url{https://arxiv.org/abs/1705.08631}

\bibitem{He_2016_CVPR}
He, K., Zhang, X., Ren, S., Sun, J.: Deep residual learning for image recognition. In: Proceedings of the IEEE Conference on Computer Vision and Pattern Recognition (CVPR) (June 2016)

\bibitem{hochreiter1997long}
Hochreiter, S., Schmidhuber, J.: Long short-term memory. Neural computation  \textbf{9}(8),  1735--1780 (1997)

\bibitem{Karpathy_2015_CVPR}
Karpathy, A., Fei-Fei, L.: Deep visual-semantic alignments for generating image descriptions. In: Proceedings of the IEEE Conference on Computer Vision and Pattern Recognition (CVPR) (June 2015)

\bibitem{klein-etal-2017-opennmt}
Klein, G., Kim, Y., Deng, Y., Senellart, J., Rush, A.: {O}pen{NMT}: Open-source toolkit for neural machine translation. In: Proceedings of {ACL} 2017, System Demonstrations. pp. 67--72. Association for Computational Linguistics, Vancouver, Canada (Jul 2017), \url{https://www.aclweb.org/anthology/P17-4012}

\bibitem{krizhevsky2012imagenet}
Krizhevsky, A., Sutskever, I., Hinton, G.E.: Imagenet classification with deep convolutional neural networks. Advances in neural information processing systems  \textbf{25} (2012)

\bibitem{lin2015microsoft}
Lin, T.Y., Maire, M., Belongie, S., Bourdev, L., Girshick, R., Hays, J., Perona, P., Ramanan, D., Zitnick, C.L., Dollár, P.: Microsoft coco: Common objects in context (2015)

\bibitem{huggingfaceSoftcatalaopennmtengcatHugging}
Mas, J.: softcatala/opennmt-eng-cat · {H}ugging {F}ace --- huggingface.co. \url{https://huggingface.co/softcatala/opennmt-eng-cat} (2022), [Accessed 08-04-2024]

\bibitem{Muller_DEW_2017}
M{\"u}ller, E., Springstein, M., Ewerth, R.: ``when was this picture taken?'' -- image date estimation in the wild. In: Jose, J.M., Hauff, C., Alt{\i}ngovde, I.S., Song, D., Albakour, D., Watt, S., Tait, J. (eds.) Advances in Information Retrieval. pp. 619--625. Springer International Publishing, Cham (2017)

\bibitem{nguyen2023show}
Nguyen, K., Biten, A.F., Mafla, A., Gomez, L., Karatzas, D.: Show, interpret and tell: entity-aware contextualised image captioning in wikipedia. In: Proceedings of the AAAI Conference on Artificial Intelligence. vol.~37, pp. 1940--1948 (2023)

\bibitem{podell2023sdxl}
Podell, D., English, Z., Lacey, K., Blattmann, A., Dockhorn, T., Müller, J., Penna, J., Rombach, R.: Sdxl: Improving latent diffusion models for high-resolution image synthesis (2023)

\bibitem{radford2021learning}
Radford, A., Kim, J.W., Hallacy, C., Ramesh, A., Goh, G., Agarwal, S., Sastry, G., Askell, A., Mishkin, P., Clark, J., Krueger, G., Sutskever, I.: Learning transferable visual models from natural language supervision (2021)

\bibitem{i2021fons}
i~Ramoneda, E.C.: El fons d’esposalles de l’arxiu de la catedral de barcelona: Q{\"u}estions i reflexions d’un usuari. Armoria. Revista d’informaci{\'o}, an{\`a}lisi i recerca de la genealogia i l’her{\`a}ldica, aix{\'\i} com tamb{\'e} de la vexil{\textperiodcentered} lologia, la sigil{\textperiodcentered} lografia, la insigni{\`a}ria, la nobili{\`a}ria i l’emblem{\`a}tica general, preferentment en l’{\`a}mbit geogr{\`a}fic i cultural catal{\`a}. (10),  33--53 (2021)

\bibitem{romero2013esposalles}
Romero, V., Forn{\'e}s, A., Serrano, N., S{\'a}nchez, J.A., Toselli, A.H., Frinken, V., Vidal, E., Llad{\'o}s, J.: The esposalles database: An ancient marriage license corpus for off-line handwriting recognition. Pattern Recognition  \textbf{46}(6),  1658--1669 (2013)

\bibitem{rumelhart1986learning}
Rumelhart, D.E., Hinton, G.E., Williams, R.J.: Learning representations by back-propagating errors. nature  \textbf{323}(6088),  533--536 (1986)

\bibitem{salem2016analyzing}
Salem, T., Workman, S., Zhai, M., Jacobs, N.: Analyzing human appearance as a cue for dating images. In: 2016 IEEE Winter Conference on Applications of Computer Vision (WACV). pp.~1--8. IEEE (2016)

\bibitem{LAION5B2022}
Schuhmann, C., Beaumont, R., Vencu, R., Gordon, C., Wightman, R., Cherti, M., Coombes, T., Katta, A., Mullis, C., Wortsman, M., Schramowski, P., Kundurthy, S., Crowson, K., Schmidt, L., Kaczmarczyk, R., Jitsev, J.: Laion-5b: An open large-scale dataset for training next generation image-text models. In: Koyejo, S., Mohamed, S., Agarwal, A., Belgrave, D., Cho, K., Oh, A. (eds.) Advances in Neural Information Processing Systems. vol.~35, pp. 25278--25294. Curran Associates, Inc. (2022), \url{https://proceedings.neurips.cc/paper_files/paper/2022/file/a1859debfb3b59d094f3504d5ebb6c25-Paper-Datasets_and_Benchmarks.pdf}

\bibitem{stacchio2023analyzing}
Stacchio, L., Angeli, A., Lisanti, G., Marfia, G.: Analyzing cultural relationships visual cues through deep learning models in a cross-dataset setting. Neural Computing and Applications pp. 1--16 (2023)

\bibitem{gemma_2024}
Team, G.: Gemma  (2024). \doi{10.34740/KAGGLE/M/3301}, \url{https://www.kaggle.com/m/3301}

\bibitem{thapliyal_crossmodal-3600_2022}
Thapliyal, A.V., Pont-Tuset, J., Chen, X., Soricut, R.: Crossmodal-3600: {A} {Massively} {Multilingual} {Multimodal} {Evaluation} {Dataset} (Oct 2022). \doi{10.48550/arXiv.2205.12522}, \url{http://arxiv.org/abs/2205.12522}, arXiv:2205.12522 [cs]

\bibitem{Vaswani_Transformer_2017}
Vaswani, A., Shazeer, N., Parmar, N., Uszkoreit, J., Jones, L., Gomez, A.N., Kaiser, L.u., Polosukhin, I.: Attention is all you need. In: Guyon, I., Luxburg, U.V., Bengio, S., Wallach, H., Fergus, R., Vishwanathan, S., Garnett, R. (eds.) Advances in Neural Information Processing Systems. vol.~30. Curran Associates, Inc. (2017), \url{https://proceedings.neurips.cc/paper_files/paper/2017/file/3f5ee243547dee91fbd053c1c4a845aa-Paper.pdf}

\bibitem{elinguistics}
{Vincent Beaufils}: elinguistics. \url{https://http://www.elinguistics.net/}, accessed: 2024-07-16

\bibitem{Vinyals_2015_CVPR}
Vinyals, O., Toshev, A., Bengio, S., Erhan, D.: Show and tell: A neural image caption generator. In: Proceedings of the IEEE Conference on Computer Vision and Pattern Recognition (CVPR) (June 2015)

\bibitem{xacquisom}
{XAC}: Què és xac? \url{https://xac.gencat.cat/ca/07_Que_es_XAC/index.html}, accessed: 2024-07-20

\bibitem{xia2019generalized}
Xia, M., Kong, X., Anastasopoulos, A., Neubig, G.: Generalized data augmentation for low-resource translation (2019)

\bibitem{yin2024surveymultimodallargelanguage}
Yin, S., Fu, C., Zhao, S., Li, K., Sun, X., Xu, T., Chen, E.: A survey on multimodal large language models (2024), \url{https://arxiv.org/abs/2306.13549}

\bibitem{zhao2021boosting}
Zhao, W., Hu, Y., Wang, H., Wu, X., Luo, J.: Boosting entity-aware image captioning with multi-modal knowledge graph (2021)

\end{thebibliography}
\end{document}